\documentclass[sigconf]{acmart}
\usepackage{tabto}
\AtBeginDocument{%
  \providecommand\BibTeX{{%
    \normalfont B\kern-0.5em{\scshape i\kern-0.25em b}\kern-0.8em\TeX}}}

\setcopyright{acmcopyright}
\copyrightyear{2023}
\acmYear{2023}

\acmConference[VAT-HRI '22]{Variable Autonomy for human-robot Teaming (VAT) workshop at the 18th ACM/IEEE International Conference on Human-Robot Interaction (HRI)}{March 13,
  2023}{Stockholm, Sweden}
\acmBooktitle{Variable Autonomy for human-robot Teaming (VAT) workshop at the 18th ACM/IEEE International Conference on Human-Robot Interaction (HRI),
 March 13, 2023, Stockholm, Sweden} 

\begin{document}


\title{A Variable Autonomy approach for an Automated Weeding Platform}

\author{Ionut Moraru}
\authornote{Both authors contributed equally to this research.}
\email{imoraru@lincoln.ac.uk}
\orcid{0000-0002-8314-037X}
\affiliation{%
  \institution{School of Computer Science, University of Lincoln}
  \city{Lincoln}
  \country{UK}}

\author{Tsvetan Zhivkov}
\authornotemark[1]
\email{tzhivkov@lincoln.ac.uk}
\affiliation{
  \institution{ Lincoln Institute for Agri-food Technology, University of Lincoln}
  \city{Lincoln}
  \country{UK}}
  
\author{Shaun Coutts}
\email{scoutts@lincoln.ac.uk}
\affiliation{
  \institution{ Lincoln Institute for Agri-food Technology, University of Lincoln}
  \city{Lincoln}
  \country{UK}}
  
\author{Dom Li}
\email{doli@lincoln.ac.uk}
\affiliation{
  \institution{ Lincoln Institute for Agri-food Technology, University of Lincoln}
  \city{Lincoln}
  \country{UK}}
  
\author{Elizabeth I Sklar}
\email{esklar@lincoln.ac.uk}
\affiliation{
  \institution{ Lincoln Institute for Agri-food Technology, University of Lincoln}
  \city{Lincoln}
  \country{UK}}


  
  
  

\renewcommand{\shortauthors}{Moraru and Zhivkov, et al.}

\begin{abstract}
  Climate change, increase in world population and the war in Ukraine have led nations such as the UK to put a larger focus on food security, while simultaneously trying to halt declines in biodiversity and reduce risks to human health posed by chemically-reliant farming practices. Achieving these goals simultaneously will require novel approaches and accelerating the deployment of Agri-Robotics from the lab and into the field. 
  In this paper we describe the \textit{ARWAC} robot platform for mechanical weeding.
  We explain why the mechanical weeding approach is beneficial compared to the use of pesticides for removing weeds from crop fields.
  Thereafter, we present the system design and processing pipeline for generating a course of action for the robot to follow, such that it removes as many weeds as possible. 
  Finally, we end by proposing a trust-based ladder of autonomy that will be used, based on the users' confidence in the robot system.  
\end{abstract}

\keywords{Robotics, Human-Robotic-Interaction, Autonomous Planning}

\begin{teaserfigure}
  \centering
  \includegraphics[width=.9\textwidth]{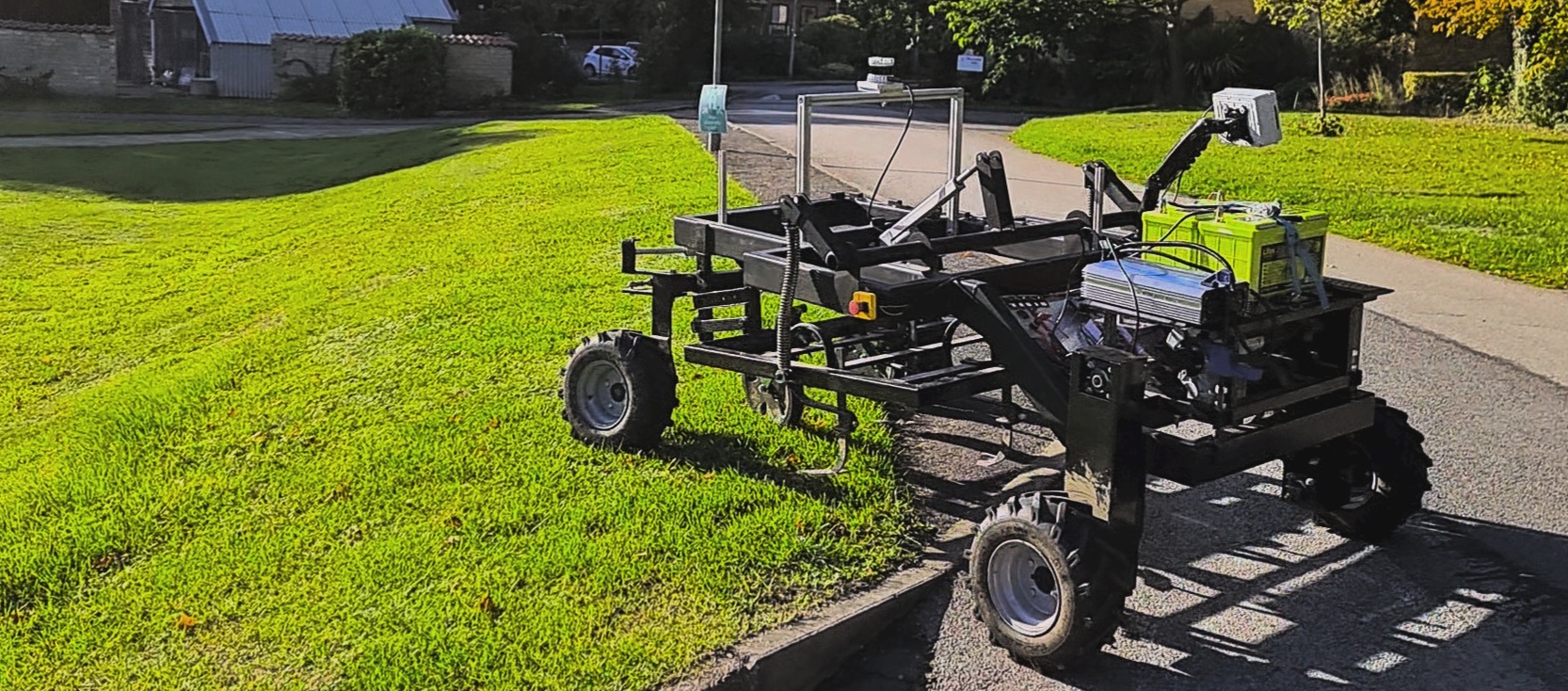}
  \caption{ARWAC Weeding Platform at Riseholme Campus, University of Lincoln. October 2022.}
  \Description{ARWAC Weeding Platform at Riseholme Campus, University of Lincoln.}
  \label{fig:v4}
\end{teaserfigure}

\maketitle

\section{Introduction}

\begin{figure*}[ht]
  \centering
  \includegraphics[width=.9\textwidth]{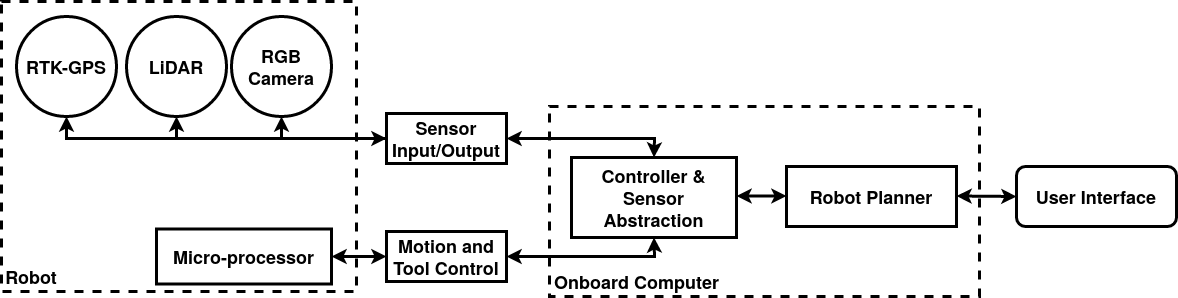}
  \caption{System design diagram of the ARWAC platform.}
  \Description{System design with regards to data flow.}
  \label{fig:system}
\end{figure*}

As the fields of AI and Robotics mature, their impact will start being felt outside of well-structured environments that have little to no outside factors modifying them. 
We take inspiration from domains such as logistics and warehouse management \cite{bogue:warehouse-robots,mahroof:ocado}, and look at how this evolution in robotics will change agriculture.

The need to feed the world's growing population as the effects of climate change accelerate, while simultaneously reducing the environmental impacts of agriculture has driven a long term focus on agri-robotics as a technology that could potentially deliver these twin goals \cite{duckett2018agricultural}. This ambition has evolved from a distant goal to a necessity in recent years, as nations feel the effects of global supply chain disruptions from the pandemic and the war in Ukraine. 
The subjects of interest in agri-robotics have been focusing on either picking soft fruits and vegetables \cite{huang2020design, navas2021soft}, precision herbicide spraying \cite{salazar2022beyond}, crop monitoring and optimisation \cite{harman2022multi}, or advancing auxiliary technology for aiding robotics in remote fields \cite{zhivkov20225g,zhivkov:5g,moraru2021using}. 

Our work focuses on non-chemical weed control. The use of critical chemical tools such as herbicides will continue to come under pressure from evolved weed resistance and tightening regulations to protect human and environmental health \cite{hicks2018factors}. 
In order to maintain global food security, both more efficient use of herbicide and non-chemical weed and pest control will be needed before we lose these valuable tools. Robotics and autonomy will be critical technologies to enable this change to occur.

The rest of the paper is comprised of three sections. Section \ref{sec2:arwac} describes the ARWAC weeding robot platform, as shown in Figure \ref{fig:v4}.
Section \ref{sec3:autonomylvls} proposes three levels of autonomy that aim at increasing the users' trust in the robot platform. 
Finally, Section \ref{sec4:related} ends with a presentation of related works from the field of Agri-Robotics, focusing on systems designed for weeding and pest removal.

\section{ARWAC Weeding Robot Platform}
\label{sec2:arwac}
The \textit{ARWAC}\footnote{ARWAC website - \url{http://arwac.uk/}} platform is a rover-type robot built specifically with the intention of creating an autonomous and carbon-neutral approach to identifying and removing blackgrass from fields without harming the cereal-type crops. 
The robot is bespoke and unique in its approach to tackle the challenge of weeding in agriculture. 
The common approach to weeding is to improve on simpler and lower probability of failure methods, such as spot-spraying.
However, mechanical weeding is more complex as it requires not only vision systems to identify weeds, but also precision tools to target the weeds.
Moreover, errors in tool displacement can cause far more damage than precision spot-spraying, therefore requiring accurate crop row detection and navigation systems.
The ARWAC robot uses a mechanical weeding tool, which removes the need for aerosols and chemicals that are bad for the environment and human health, and can damage and kill crops.

The novelty behind the ARWAC robot is that it was designed to be relatively light for agricultural tasks, with a weight under 250 kg, as this will reduce soil compaction, increase battery life and can lead to the robot being able to do more passes in a season under a wider range of ground conditions, increasing the effectiveness of mechanical weeding.
Moreover, it is a low cost platform utilising a low power micro-processor for driving the motors and Ackermann steering, and an onboard computer to perform sensor fusion and autonomous planning.

\subsection{System Design}
The system design of the robot consists of a central computer that receives data from an ESP32\footnote{ESP32 - \url{https://docs.espressif.com/projects/esp-idf/en/latest/esp32/}} micro-processor, Ouster OS1\footnote{Ouster OS1 - \url{https://data.ouster.io/downloads/hardware-user-manual/hardware-user-manual-revd-os1.pdf}} LiDAR, RGB camera and Ardusimple RTK-GPS (sensors), this is illustrated in Figure \ref{fig:system}.
It uses the data from the micro-processor to determine if the motors are functional and if the robot is moving, whereas inputs from the LiDAR, RGB and RTK-GPS are used to determine a safe route to navigate in the environment and identify weeds in the field.
The central computer uses data from the sensors to send control commands back to the micro-processor to stop, move or avoid collisions.
Additionally, the sensor data is also used to send control commands to the micro-processor to control the weeding tool.
It should be noted that the central computer is using ROS1 (Robot Operating System \cite{quigley:ros}) for communication and control of sensors and robot motion.
 
 \begin{figure*}[t]
     \centering
     \includegraphics[width=0.85\linewidth]{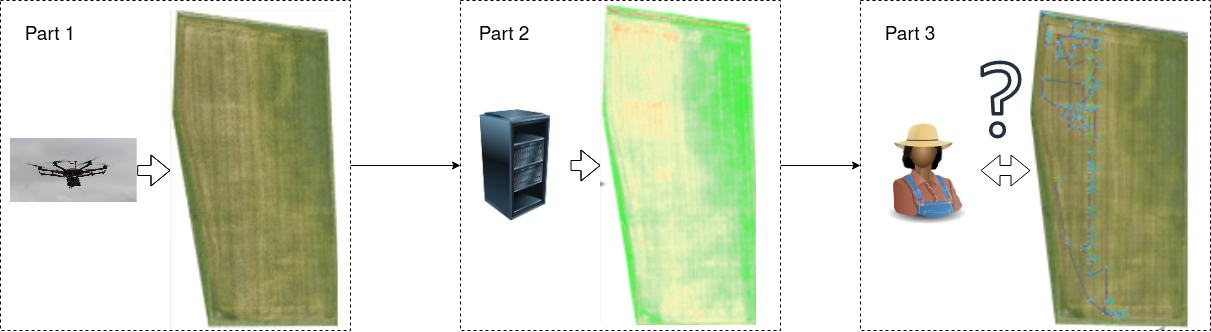}
     \caption{Three part pipeline -- mapping, processing and planning -- for generating a sequence of actions that the ARWAC Platform uses for mechanically removing the blackgrass from the field of crops.  }
     \label{fig:Vat_pipeline}
 \end{figure*}

The control code is abstracted from the planner, 
allowing the planner to be used as the main interface to control the robot's behaviour, i.e., set, remove, stop and edit robot tasks.
The user (farmer) will be able to engage with the robot via a simplified mobile app or GUI, made to directly communicate with the planner. 
Rosbridge\footnote{rosbridge\_suite package - \url{http://wiki.ros.org/rosbridge_suite}} is a ROS1 package that provides an easy to use interface for any application, specifically web-based, to interact and control ROS-based applications.
This is what we will use to allow the user to interact with the robot and configure the desired level of autonomy, with basic descriptions of each level.



\subsection{System Processing Pipeline}

The project's proposed processing pipeline can be seen in Figure \ref{fig:Vat_pipeline} and is split in three parts: mapping, pre-processing and planning. The first one consists of a drone being flown over the field of crops, taking pictures which will then be stitched together as a map. This will then be uploaded to a central computer which will generate a probability weed map, which will inform the system of how likely a spot is to have blackgrass growing in it. This step could also be done with the input of farmers with a suitable interface to allow them to input their knowledge of weed distribution on their fields to a digital map.

The final step consists of synthesising a plan based on the areas identified as most likely to have blackgrass growing around the arable crop. This part is vital, as it will offer the agent high-level reasoning process. This is based on PDDL planning \cite{fox2003pddl2, moraru2019simplifying}, where solvers create a plan based on the actions the agent poses to modify the agent such that at the end, the environment will look as desired by the user. For this implementation, there will be the possibility of adapting the plan based on the users input, where we have three approaches based on users trust in the systems reasoning process.

\section{Trust-based Autonomy Levels}
\label{sec3:autonomylvls}
Safe and trusted/explainable AI is one of the fastest growing sub-fields in computer science research, due to many more AI approaches being incorporated into systems currently being in production. In Ref.~\cite{jacovi2021formalizing}, the notion of trust in AI is formalised from many different perspectives, and we choose to work with the view of "trust in model correctness", which they describe as being not how well the approach works, but how easy it is for the user to distinguish where the AI succeeds or fails. 

In a perfect world, where the user trusts the robot platform, it would be able to act completely independent from the user and carry out its plan without any need of validation or verification from outside. However, agri-robotic platforms are still far from perfect and cause considerable economical and physical harm to the crops or the user, with the ARWAC robot having the same constraints. For such a system to be broadly adopted by the agricultural sector, having the users' trust will be vital to be earned, as shown in \cite{bedue2021can,canal2020building}.


For users to better understand and learn to trust the capabilities of the ARWAC weeding platform, we have encorporated the robot with three levels of autonomy for high-level reasoning. Their intended aim is so they would lead to gaining the users trust in the platforms's ability to complete its mission in a safe and efficient manner, while also being able to verify that there are no errors in the processing pipeline for their specific farm. 

The three levels of autonomy are: 
\begin{itemize}
    \item Low-Trust Autonomy -- the user will come up with a plan for the platform to follow, action by action, ;
    \item Partial-Trust Autonomy -- the robot comes up with a plan which it proposes to the user \cite{canal2022planverb}, who will then challenge it until they are satisfied with it, or reject and come up with a new one \cite{krarup2021contrastive};
    \item Full-Trust Autonomy -- it synthesises a plan which it will follow without the need for human validation.
\end{itemize}

These three levels would allow time for the users to fully understand the capabilities and limitations of the ARWAC weeding platform. It will do this by first by first operating in a fully manual mode entitled "Low-Trust autonomy", made with the intention that the user would pay closer attention to the platform and which would lead to the user learning how the platform would operate specifically on their farm. This mode is intended to be entered initially, and used until the user is familiar with the robot and has used the platform on a representative number of their fields.

Following this mode, the second level of autonomy would be the "Partial-Trust autonomy", is designed for familiarising the user with platforms planning approach to weeding a farm. This would be done by having the robot propose a plan of action to the user, who would then be able to challenge it by asking for contrastive explanations \cite{krarup2021contrastive}. 

Finally, once the user is accustomed to how the platform would tackle weeding, from their individual actions to their own plans, the system has a "Full-Trust Autonomy" mode, which would allow for the ARWAC weeding platform to fully autonomously complete a user selected field without any human intervention.

Our current schedule is to evaluate the capabilities of the ARWAC robot during the spring season (late February to early May 2023), leading into studies of users' direct interaction with it. Once the platform is fully validated, we will start evaluating our proposed autonomy levels such that we ensure obvious bugs will not play the main role in how users will build trust in the system.

\section{Related Works}
\label{sec4:related}
Outdoor robotic activities range from agricultural\footnote{Agri-robotics - \url{https://sagarobotics.com/}, \hspace{0.15cm} \url{https://www.smallrobotcompany.com/}}, delivery\footnote{Robot Delivery - \url{https://www.starship.xyz/}} to autonomous shuttling\footnote{Robot Shuttle - \url{https://www.navya.tech/en}}. Generally, outdoor environments in this field are commonly referred to as unstructured or semi structured. 
There are two main reasons for this.
Firstly, in outdoor environments there are few or no permanent obstacles/walls that can be used to memorise (build) a map, which can be used to localise a robot with. 
Secondly, communication and control of dynamic events cannot be guaranteed. For example, in an indoor environment, navigating around people is the dynamic event, and communication can be guaranteed to a certain extent, with predicted behaviour and fallback methods. 
Comparatively, in an outdoor environment, dynamic events are unbounded (weather, animals, vehicles, etc.,) and communication cannot be guaranteed.
This makes outdoor navigation and localisation more complex than it is for indoor, and GPS data is a necessity \cite{ohno:gps}.
However, there are multiple methods that can be used to improve sensor fusion and thereby improve outdoor localisation \cite{mahony:place-recognision}; e.g. recognition/horizon detection.

Agri-robotic weeding platforms exist in two main forms, herbicide sprayers and mechanical weeders.
Both forms fall under the umbrella term of precision agriculture (PA). 
Moreover, both forms generally make use of machine vision systems to identify the exact location of weeds \cite{li:ml-weeding,salazar:spot-spraying,salazar2022beyond}, which allow the sprayer or weeding tool to carry out its function.
Mechanical weeders require combinations of weeding and cultivation strategies to achieve acceptable weed control, compared to their sprayer counterpart \cite{machleb:mech}. 
There are multiple tool mechanism that are used by mechanical weeders such as harrow \cite{bakker:harrow-hoe} and rotating \cite{norremark:cycloid-hoe} tools.
The ARWAC robot makes use of a harrow-like mechanism with blades that cut the identified weeds.

\begin{acks}
We would like to express our gratitude toward ARWAC for providing their robot platform and their continued help toward the project. Also to the farmers whose continued help and input has been and will be critical to the success of this project. This work is supported by Innovate UK (UKRI) under grant no. \#10027958 and project titled ``\textit{ARWAC attack Blackgrass in Farming}''.
\end{acks}

\bibliographystyle{ACM-Reference-Format}
\bibliography{main}

\end{document}